\documentclass[lettersize,journal]{IEEEtran}
\usepackage{amsmath,amsfonts}
\usepackage{algorithmic}
\usepackage{array}
\usepackage[caption=false,font=normalsize,labelfont=sf,textfont=sf]{subfig}
\usepackage{textcomp}
\usepackage{stfloats}
\usepackage{url}
\usepackage{verbatim}
\usepackage{graphicx}
\usepackage{soul}
\usepackage{url}
\usepackage[hidelinks]{hyperref}
\usepackage[utf8]{inputenc}
\usepackage[small]{caption}
\usepackage{booktabs}
\usepackage{times}
\usepackage{cite}
\usepackage{algorithm}
\usepackage{color}
\usepackage{multirow}
\usepackage{colortbl}
\usepackage{amssymb}
\usepackage{amsthm}
\usepackage{hyperref}
\hypersetup{
    colorlinks=true,
    linkcolor=blue,
    filecolor=blue,      
    urlcolor=blue,
    citecolor=black,
}

\newcommand{\Lily}[1]{{\color{black}#1}}

\def\BibTeX{{\rm B\kern-.05em{\sc i\kern-.025em b}\kern-.08em
    T\kern-.1667em\lower.7ex\hbox{E}\kern-.125emX}}
\usepackage{balance}

\begin{document}
\title{MIFI: \textbf{M}ult\textbf{I}-camera \textbf{F}eature \textbf{I}ntegration for Roust 3D Distracted Driver Activity Recognition}

\author{Jian Kuang, Wenjing Li, Fang Li, Jun Zhang, Zhongcheng Wu
\thanks{Manuscript created 31 July 2022; revised
 16 May 2023; accepted 7 August
2023.
This work was supported by the Hefei Comprehensive National Science Center through the Pre-research Project on Key Technologies of Integrated Experimental Facilities of Steady High Magnetic Field and Optical Spectroscopy and the High Magnetic Field Laboratory of Anhui Province.
The Associate Editor for this article was S. A. Birrell.
(Jian Kuang and Wenjing Li contributed
equally to this work.)
 (\textit{Corresponding Author: Wenjing Li, Jun Zhang}.)

Jian Kuang, Wenjing Li,  Fang Li, Jun Zhang, and Zhongcheng Wu are with High Magnetic Field Laboratory, HFIPS, Chinese Academy of Sciences, Hefei, China, and University of Science and Technology of China, Hefei, China. They are also with the High Magnetic Field Laboratory of Anhui Province, Hefei, China. (email: wjli007@mail.ustc.edu.cn)

}}

\markboth{Journal of \LaTeX\ Class Files,~Vol.~18, No.~9, AUGUST~2023}%
{How to Use the IEEEtran \LaTeX \ Templates}

\maketitle

\begin{abstract}

Distracted driver activity recognition plays a critical role in risk aversion-particularly beneficial in intelligent transportation systems.
However, most existing methods make use of only the video from a single view and the difficulty-inconsistent issue is neglected.
Different from them, in this work, we propose a novel \textbf{M}ult\textbf{I}-camera \textbf{F}eature \textbf{I}ntegration (MIFI) approach for 3D distracted driver activity recognition by jointly modeling the data from different camera views and explicitly re-weighting examples based on their degree of difficulty. 
Our contributions are two-fold: (1) We propose a simple but effective multi-camera feature integration framework and provide three types of feature fusion techniques.
(2) To address the difficulty-inconsistent problem in distracted driver activity recognition, a periodic learning method, named example re-weighting that can jointly learn the easy and hard samples, is presented.
The experimental results on the 3MDAD dataset demonstrate that the proposed MIFI can consistently boost performance compared to single-view models. The source code is available at \href{https://github.com/john828/MIFI}{https://github.com/john828/MIFI.}

\end{abstract}


\section{Introduction}
\IEEEPARstart{C}ar accident has been one of the biggest worldwide killers, leading to 1.35 million deaths every year. The report from World Health Organization (WHO) shows that car accident has been the leading cause of death among children and young people aged 5 to 29 \cite{world2018global, nemcova2021multimodal}. According to research conducted by Volvo, almost 90\% of traffic accidents are caused by drivers, which can be attributed to factors such as distracted driving, misjudgment of risks, and other related factors \cite{kockum2017volvo}. 
Distraction behaviors take the driver’s attention away from driving to the secondary task such as \textit{sending a text message}, \textit{talking on a cell phone}, which can greatly increase the chance of car crash\cite{azadani2021driving, wang2021survey}.

\textbf{D}istracted \textbf{D}river \textbf{C}lassification (DDC) can be considered a subset of general action recognition, an area that has made impressive progress thanks to recent advances in deep neural networks \cite{simonyan2014two, carreira2017quo, fang2021dada}. However, DDC presents two unique challenges that distinguish it from general action recognition. Firstly, DDC is a fine-grained action recognition task, as distracted behaviors are executed by the same subject with similar body movements, resulting in subtle differences \cite{wharton2021coarse}. For instance, the only difference between \textit{texting left} and \textit{talking left} is the movement of the left arm. Secondly, the scene context in DDC cannot be used because the background of different distracted behaviors is the same. This contrasts with general action recognition, where specific scene contexts, such as \textit{playing basketball on a basketball court} or \textit{swimming in a pool}, can be utilized to quickly recognize activities. Consequently, fine-grained differences and the scarcity of scene context information make DDC much more challenging than general action recognition.
\begin{figure}[t!]
\centering
    \includegraphics[width=\linewidth]{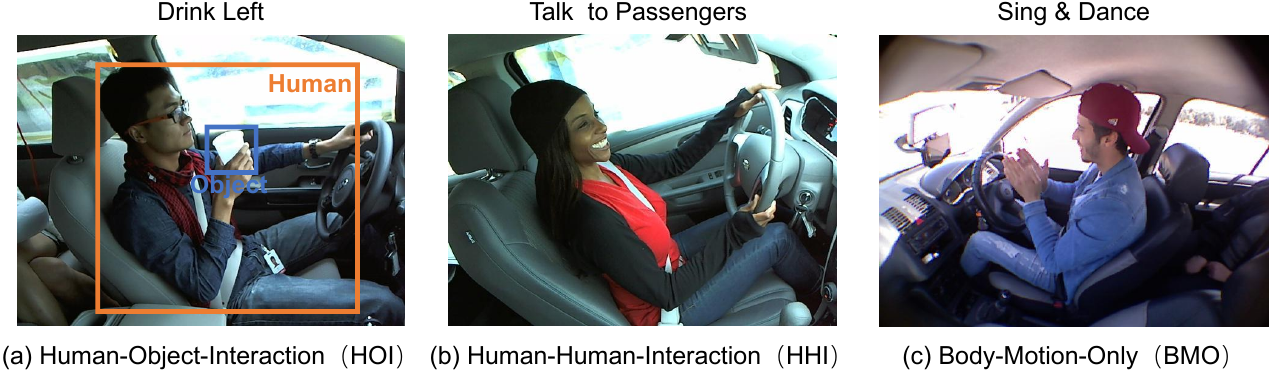}
    \caption{Three types of distracted behaviors.}
    \label{fig:three-type}
    \vspace{-6mm}
\end{figure}

Due to its application and research
significance, distracted driver classification (DDC)  has become increasingly popular in the community.
Generally, DDC models can be divided into two lines, image-based \cite{abouelnaga2017real, xing2019driver, baheti2018detection}
and video-based models \cite{jegham2019mdad, billah2018recognizing}.
Image-based methods recognize distracted behaviors through a single image, which has received competitive performance in identifying Human-Object-Interaction (HOI)
type of behaviors where an iconic object exists, for example, the bottle in drinking  (Figure~\ref{fig:three-type}(a)), the phone in talking.
But the image-based approaches often fail to distinguish the Human-Human-Interaction (HHI) (Figure~\ref{fig:three-type}(b)) and  Body-Motion-Only (BMO) (Figure~\ref{fig:three-type}(c)) types of activities due to the limited information in a single image.
To address this issue, researchers employ video-based \textit{i.e.,} 3D techniques~ \cite{lin2019tsm, wang2021action} to capture spatial-temporal relationships among frames. 
Despite its high computational load, 3D models are proven to be more suitable for accuracy-priority tasks~ \cite{hara2018can} including DDC.
Consequently, 3D models~ \cite{zhou2018temporal, wang2021action, lin2019tsm, hu2018squeeze} have emerged in the research community.

\begin{figure}[t!]
    \centering
    \includegraphics[width=\linewidth]{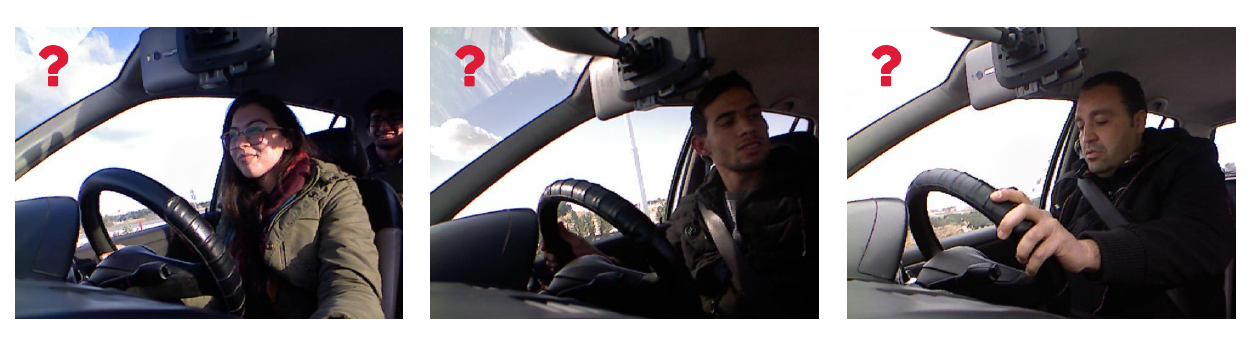}
    \caption{The information from a single-view video is often limited, making us hardly recognize the distraction.}
    \label{fig:example-one-view}
    \vspace{-6mm}
\end{figure}

Although the performance of video-based approaches has rapidly
increased, there are several problems that still largely remain unsolved in the field.
\textit{First}, current models only focus on single-view-based DDC, making them hard to capture some critical features due to the limited radiation range of a single camera.
Some examples are shown in Figure \ref{fig:example-one-view},
it is also difficult for us humans to recognize the behaviors from a limited perspective.
\textit{Second}, the difficulty imbalanced problem is neglected.
As shown in Figure \ref{fig:three-type}, distracted driver activities are made up of different types of behaviors where the HOI type of behavior is much easier than the HHI and BMO type of activities.
The diversity in terms of behavior type will lead to the difficulty inconsistent, \textit{i.e.,} there exists \textit{easy} and \textit{hard} samples in DDC. Thus, directly using average loss to update the models will lead them to fail to recognize hard samples.

To address the aforementioned issues, we propose a novel framework for DDC. To tackle the first issue, we introduce a multi-view feature learning technique that uses video from multiple cameras, which has two notable advantages. First, the features captured from different cameras can complement each other and improve the overall accuracy of the model. Second, onboard cameras are relatively inexpensive and easy to deploy. To address the second concern, we present a sample re-weighting strategy that assigns losses dynamically to samples during back-propagation. In summary, our contribution can be three-fold:

\begin{itemize}

 \item
We carefully analyze the properties and difficulties of DDC,
and identify the issue of limited perspective in single-view models and difficulty inconsistent in distracted classes,
helps us better understand and solve this problem.

 \item
To address the limited perspective problem in single-view models, 
we propose a novel \textbf{M}ult\textbf{I}-camera \textbf{F}eature \textbf{I}ntegration (MIFI) leveraging the information from multiple cameras, which significantly improves the performance.

\item

To address the difficulty inconsistent in distracted classes, we present a simple but effective sample re-weighting approach to encourage the model learning effectively.

\end{itemize}

\section{Related work}

In this section, we briefly review these two kinds of driver distraction classification approaches, \textit{i.e.,} image-based and video-based methods.
In addition, we also will discuss some re-weighting related works.

\subsection{Image-based Driver Distraction Classification}

Image-based DDC algorithms are boosted by the introduction of 
image-based DDC datasets, such as StateFarm and AUC \cite{farm2016state, abouelnaga2017real}.
The image-based DDC methods can be roughly divided into two streams, local-body-clue-based approaches, and global-posture-based techniques.

\textit{Local-body-clue-based approaches} often try to recognize distractions by analyzing the specific body part, such as hand, face, eyes, and \textit{etc}.
For example, Yan \textit{et al.} \cite{yan2016driving} propose to identify 
four types of behaviors including normal safe driving, operating the
shift gear, eating or smoking, and talking with a cell phone by monitoring the driver's hand position.
But the coverage of local-body-clue-based approaches is generally limited.
Thus, the researchers begin to recognize driving behavior by combining the characteristics of different parts of the body.
Yehya Abouelnaga \textit{et al.} \cite{abouelnaga2017real} classify ten types of driver behavior by hand and facial features.
And Craye \textit{et al.} \cite{craye2015driver} use four sub-modules to extract the driver's eye behavior (detection of gaze and blink), arm position (right arm up, down, forward, and right), head direction and facial expression.
To evaluate which part of the driver's body is of more importance to classification, Xing \textit{et al.} \cite{xing2017identification} analysis the impact of different visual elements on the classification results by combining different visual elements of drivers.

\textit{Global-posture-based techniques} identify the distracted behaviors by the driver's posture, which is able to cover a more comprehensive set of distracting actions \cite{li2022learning}.
In order to alleviate the influence of the background information,
Yang \textit{et al.} \cite{xing2019driver} use the Gaussian mixture model to segment the original image and separate the driver's body from the image background.
Lei \textit{et al.} \cite{zhao2021driver} cascade multiple attention-based convolutional neural networks to realize the extraction of the adaptive discrimination spatial region of the driver image.
Chen \textit{et al.} \cite{huang2020hcf} combine resnet50, inception V3, and xception to extract driver behavior features based on migration learning, and connect the extracted features to make better use of the driver's overall characteristics.

Although the image-based approaches have achieved competitive performance in some public datasets, they often failed to identify the Human-Human-Interaction and Body-Motion-Only types of distractions such as \textit{talking to passengers}, \textit{singing}, which are also very dangerous in piratical. For example, talking to passengers may lead the drivers' attention away from the road.
This will result in a narrow range of distracted diver classification applications.

\subsection{Video-based Driver Distraction Classification}

Video-based driver distraction classification has recently demonstrated great potential in learning spatial-temporal relations between frames
followed with the progress in general 3D-based action recognition task \cite{tran2015learning,carreira2017quo,feichtenhofer2020x3d}.
These methods could be categorized into two types: (1) multi-modality
method often takes two inputs of RGB and optical flow to separately model appearance and motion information in videos with a late fusion.
For instance, Chen \textit{et al.} \cite{chen2020driver} present a 
two-stream network \cite{simonyan2014two} taking the optical flow and RGB features as two separate inputs to classify driver behaviors.
The work \cite{behera2018context} proposes a two-stream M-LSTM  focusing on appearance information with two different levels of abstraction.
However, extracting optical flow features is time and energy-consuming.
Current methods \cite{wang2018temporal} often first obtain and save the optical flow feature, then directly use it as input to do the inference.
This makes these kinds of methods only suitable for video classification problems where the videos have been completely obtained, while our focus in this paper is on the driver’s ongoing activity from partial observation \cite{behera2020deep}. 
(2) single-modality approach takes only the RGB data as input and tries to represent the spatial-temporal relations from the video.
Xing \textit{et al.} \cite{xing2021multi} propose to identify
whether the driver's behavior is dangerous or not by estimating the driver's state through the LSTM network.
Pan \textit{et al.} \cite{pan2021driver} obtain the driver's motion state information through LSTM and combined it with the spatial information in the frame.
Tan \textit{et al.} \cite{tan2021bidirectional} use two convolutional neural networks to extract the appearance and pose features in the video.  And the complementary information between the two features is used to make the model more focused on the key local area of the video frame related to the driver's behavior.
Although single-modality methods have achieved gratifying results, they are limited in recognizing some distractions where a large range of motion exists such as operating radio, reaching behind
because of the limited range of the camera.
In this work, we argue that multiple cameras may compensate each other to obtain higher accuracy.
Therefore, we propose a novel multi-view feature learning framework 
taking the videos from multiple cameras installed at
different locations, which can greatly facilitate performance.
\begin{figure*}[t!]
    \centering
    \includegraphics[width=\linewidth]{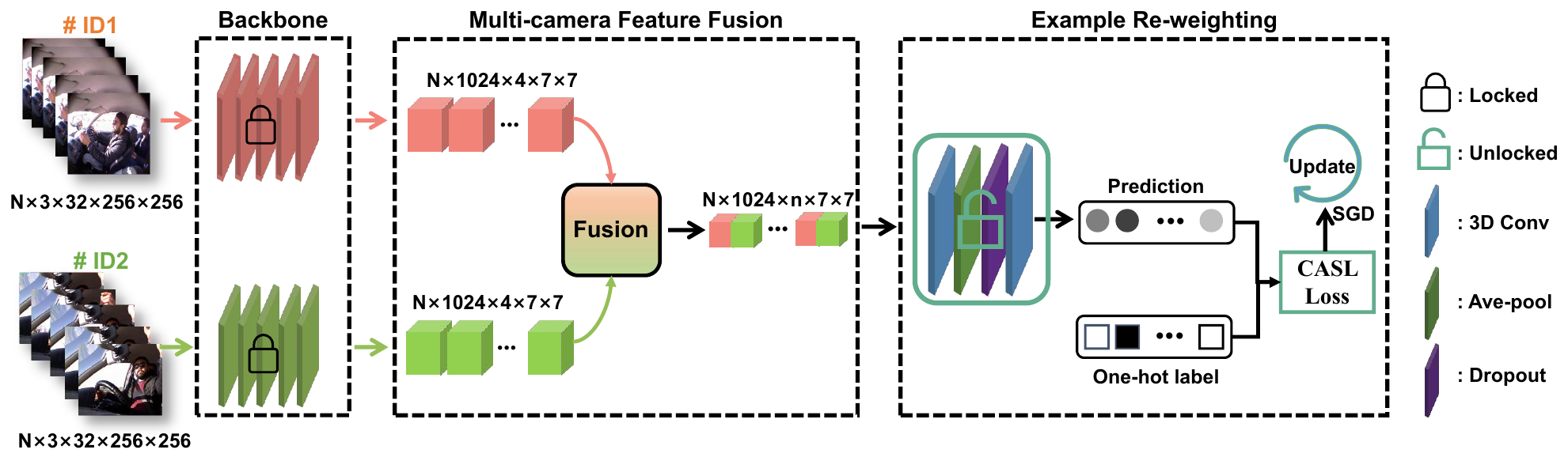}
    \caption{The framework of the proposed MIFI method.}
    \label{fig:framework}
    \vspace{-6mm}
\end{figure*}

\vspace{-2mm}
\subsection{Example reweighing}

The idea of example re-weighting has been studied from very previous to recently due to its superiority in dealing with the training set bias phenomenon.
Previous approaches can be traced back to importance sampling \cite{kahn1953methods} and boosting models \cite{freund1997decision}.
Recently,
Cui \textit{et al.} \cite{cui2019class} proposes a method to calculate the effective sample size of a class and balances the loss contribution of each class by the effective sample size.
Ren \textit{et al.} \cite{ren2018learning}propose a novel meta-learning algorithm that learns to assign weights to training examples based on their gradient directions.
More recently, some popular loss functions have been proposed, like Focal Loss (FL) \cite{lin2017focal}, Asymmetric Loss (ASL) \cite{ben2020asymmetric}, Balanced MSE \cite{ren2022balanced} and \textit{etc}, which has been demonstrated that they can down-weight
easy examples and thus focus training on hard negatives.
Unfortunately, these works mainly focus on imbalanced data whereas the data distribution is relatively balanced in the DDC task.
Thus, in this work, we present a novel example re-weighting method to address the difficult inconsistent but data-balanced problem.

\section{Methodology}

We start by describing the problem definition in Sec \ref{sec:problem-def}, followed by the overview of the proposed method.
In Sec \ref{sec:single}, we describe the training strategy of the single-view backbone.
In Sec \ref{sec:fusion}, we present the multi-camera integration model, that exploits the features of multiple cameras.
In Sec \ref{sec:re-weighting}, the example re-weighting technique is introduced that is able to address the difficulty-inconsistent phenomenon in DDC.

\subsection{Problem definition and overview of MIFI}
\label{sec:problem-def}

Multi-camera driver distraction recognition task tries to classify distracted behaviors by multiple inputs.
Intuitively, different views usually contain complementary information, and joint learning of multiple cameras is able to learn a quite effective feature for DDC.
For simplicity, we consider a two-camera scenario.
Each sample can be denoted as $x_{i} = \{(x_{i}^{1}, x_{i}^{2}), y_{i}\}$ where $x_{i}^{1}$, $x_{i}^{2} \in R^{3\times T\times 224 \times 224}$ are the data from camera \#ID 1 and camera \#ID 2, $T$ is the number of frames, and $y_{i}$ is the corresponding label.
Our goal is to use $x_{i}^{1}$ and $x_{i}^{2}$ to predict the label of such sample.
It is quite obvious that the data distribution from different cameras varies greatly.
\Lily{The challenge consists of how to fusion the feature of different cameras}.
In this paper, we propose a simple yet effective multi-camera integration strategy, called MIFI, to connect the data from different views to boost the performance of DDC.
The framework of the proposed method is shown in Figure \ref{fig:framework}.
In contrast to using the shared backbone, we propose to fuse the two cameras after feature representation.
Thus, we first train the two backbones separately and use them as feature extractors for the two views of inputs.
Then, the two views of features are integrated by fusion strategies including sum and concatenation fusion.
Lastly, the integrated feature will feed into the classifier layer to get the predicted label.
And we propose to use a novel cyclical focal loss by which the difficulty-inconsistent problem can be carefully addressed to update the parameters in the classifier layer while the parameters in the backbones will be fixed from both an efficient and effective perspective.

\subsection{Single-view Training }
\label{sec:single}
We first train the classification model for the data from different cameras separately.
Each model contains two parts, \textit{i.e.,} the feature extractor, denoted as $f (\cdot; \theta)$ where $\theta$ is the learned parameters, followed by the classifier $g(\cdot; \varphi)$ where $\varphi$ is the learned parameters.
We feed the input $x_{i}^{j}$ to the feature extractor  $f (\cdot; \theta)$ to obtain the feature representation of the input,

\Lily{
\begin{equation}
\widetilde{x}_{i}^{j} = f(x_{i}^{j};\theta_{j} ),
\quad
\widetilde{x}_{i}^{j} \in R^{1024\times 4\times 7\times 7},
\end{equation}
where $j$ represents the $j^{th}$ camera, and $\{1024\times 4\times 7\times 7\}$ represent $\{channel, temporal, 
 width, height\}$ dimension.
Then the predicted label can  be obtained by,

\begin{equation}
\widetilde{y}_{i}^{j}  = g(\widetilde{x}_{i}^{j} ),
\quad
\widetilde{y}_{i}^{j} \in R^{1\times 16},
\end{equation}
where $j$ represents the $j^{th}$ camera, and dimension 16 represents the number of distraction classes.
}

The parameter can be updated by minimizing the cross-entropy loss, 

\begin{equation}
Loss = -\sum_{i =0}^{n}y_{i} \log_{}{\widetilde{y}_{i}},
\end{equation}
where $n$ is the number of classes.

\subsection{Multi-view Integration}
\label{sec:fusion}
The key to multi-view integration is where and how to integrate the information from the multiple cameras. Thus, we will discuss early and later fusion strategies and then three types of fusion methods will be presented.

\noindent\textbf{Early vs. Later Fusion}
According to the definition,
early fusion integrates features before learning concepts while
later fusion first learns feature representation and then uses them to learn new concepts \cite{snoek2005early}. 
Previous literature demonstrates that later fusion is good at dealing with multi-modal data fusion like text \& image, whereas early fusion may be skilled at fusing the same modality data \cite{gunes2005affect, snoek2005early}.
\Lily{
In this work, even though the multi-view videos are the same modality, we argue that later fusion is better for multi-camera data fusion for one reason at the feature level. 
DDC is a fine-grained action recognition task and different views of the video are captured from the same vehicle with the same background leading to the same shallow features (like color, and texture).
Directly fusing shallow features  \cite{zhang2021survey}, 
will make it hard to be beneficial to the model.
In contrast, the deep features can represent high-level semantic information such as the hand, and phone \cite{zhang2021survey}, which is more critical for classification.}
Thus, later fusion may be the better choice for this task, and
we will verify this in the experiments.

Next, we discuss the multi-camera fusion method.
There are three types of fusion methods, \textit{i.e.,} sum fusion, channel concatenation fusion, and temporal concatenation.

\noindent\textbf{Sum fusion}
The features from different views can be fused by sum operation.
To be simple, let us directly take the feature produced by feature extractors, ${x}_{i}^{1}$ from camera \#ID 1, ${x}_{i}^{2}$  form camera \#ID2 as input, and the fused feature can be obtained by,

\begin{equation}
\hat{x}_{i}^{f}=\widetilde{x}_{i}^{1} + \widetilde{x}_{i}^{2},
\end{equation}
where $\hat{x}_{i}^{f}, \widetilde{x}_{i}^{1}, \widetilde{x}_{i}^{2}\in R^{C\times T\times W\times H} $. Sum fusion will not increase the learned parameters of the downstream layers.

\noindent\textbf{Channel Concatenation}
Channel concatenation stacks the features from multiple cameras in the channel dimension. The fused feature can be obtained by,

\begin{equation}
\hat{x}_{i}^{f} = CAT(\widetilde{x}_{i}^{1}, \widetilde{x}_{i}^{2}, dim= C),
\end{equation}
where CAT is the concatenation operation, $\widetilde{x}_{i}^{1}, \widetilde{x}_{i}^{2}\in R^{C\times T\times W\times H}$ and $\hat{x}_{i}^{f} \in R^{2C\times T\times W\times H}$.

\noindent\textbf{Temporal Concatenation}
Similarly, temporal concatenation fuses the features from the temporal dimension. The fused feature can be obtained by,

\begin{equation}
\hat{x}_{i}^{f} = CAT(\widetilde{x}_{i}^{1}, \widetilde{x}_{i}^{2}, dim = T),
\end{equation}
where $\widetilde{x}_{i}^{1}, \widetilde{x}_{i}^{2}\in R^{C\times T\times W\times H}$ and $\hat{x}_{i}^{f} \in R^{C\times 2T\times W\times H}$.
Channel concatenation and temporal concatenation will increase the number of features since the feature dimension is increased after the concatenation operation.

\Lily{
The key difference between channel concatenation and temporal concatenation is in terms of the concatenation dimension. Channel concatenation involves concatenating the features from two views along the channel dimension, while temporal concatenation involves concatenating the features along the temporal dimension. In other words, channel concatenation combines the features from different channels at the same time step, which can be useful for capturing spatial information across different modalities. And temporal concatenation combines the features from the same channel at different time steps,  which can be useful for capturing temporal changes over time. 
}

\subsection{Example Re-weighting}
\label{sec:re-weighting}

The motivation for example re-weighting is due to the difficult inconsistent phenomenon in DDC.
Directly minimizing the cross-entropy loss will cause lacking learning for some hard classes.
Difficulty inconsistent problem has attracted considerable attention in which Focal Loss \cite{lin2017focal}, Asymmetric Loss (ASL) \cite{ben2020asymmetric} are the most representative works.
But they all focus on the class-imbalanced task, while the class distribution in the DDC task is relatively balanced.
Thus, inspired by \cite{smith2022cyclical}, we proposed to use the Cyclical Focal Loss (CASL) to re-weight samples.
Firstly, the fused feature $\hat{x}_{i}^{f}$ is fed into the classifier layer to obtain the probability, 

\begin{equation}
p  = g (\hat{x}_{i}^{f};\varphi),
\end{equation}

Where $p \in R^{1\times n}$, $n$ is the number of classes, $\varphi$ is the parameter of the classifier $g(\cdot)$.
Then the predicted $\hat{y}_{i}$ can be utilized to compute the CASL loss which adopts a cyclic method to encourage the model learning effectively by shifting the focus between easy and hard samples.
To be specific, CASL loss, denoted as $L_{casl}$, consists of two terms, \textit{i.e.,} \textit{$L_{e}$} and \textit{$L_{h}$}, the former keep focusing on the easy samples while the latter is responsible for the hard samples,

\begin{equation}
L_{casl} = \alpha L_{e} + (1-\alpha)L_{h},
\label{Loss}
\end{equation}

Where $\alpha$ is an important hyper-parameter controlling the contribution of the two terms at each stage.
When $\alpha > 0.5 $, the model will pay
more attention to easy samples. On the contrary, it will 
care more about the hardy samples. In practice, we set $\alpha$ starting from a relatively small value,  encouraging learning easier examples first. And we use a piece-wise function to generate the value of $\alpha$,

\begin{equation}
\alpha = \begin{cases}
  1-\beta \times\frac{e_{i}}{e_{t}} & \text{ if } \beta \times e_{i}\le e_{t} \\
  ( \beta \times\frac{e_{i}}{e_{t}}-1 )/(\beta-1) & \text{ if } \beta\times e_{i}> e_{t}
\end{cases} 
\label{omega define}
\end{equation}

Where \textit{$\beta$} is cyclical factor,
\textit{$e_{i}$} is the current training epoch number and \textit{$e_{t}$} is the total training epochs.

\Lily{And $L_{e}$ can be computed by,

\begin{equation}
L_{e} = -(1+p)^{\gamma}log(p).
\label{Lp}
\end{equation}
When a sample is well-classified, \textit{i.e.,} $p\to 1$, the loss will be enlarged by the modulating factor $(1+p)^{\gamma}$ if the parameter $\gamma>1$.
This will lead to a more confident
training sample weighted more heavily 
where the parameter $\gamma$ is to control the rate at which the easy samples will be up-weighted.
If $\gamma = 0$, the $L_{e}$ loss is equivalent to cross-entropy loss.
And $L_{h}$ can be computed by,
\begin{equation}
L_{h} = -(1-p)^{\lambda _{1}}log(p) -(p)^{\lambda_{2}}log(1-p).
\label{Ln}
\end{equation}

Note that simply using the first term, \textit{i.e.,} the focal loss can lead to the model focusing on learning features from hard samples while neglecting learning features from easy samples.
Thus, we introduce the second term to keep the model focus on hard samples while maintaining the contribution of easy samples.
The parameter $\lambda _{1}$ and $\lambda _{2}$ determine the contributions of hard and easy samples.
As shown in Fig~\ref{fig:paremeter}, the bigger the $\lambda _{1}$, the fewer contributions the hard samples, and the bigger the $\lambda _{2}$, the fewer contributions the easy samples.

\begin{figure}[t!]
    \centering
    \includegraphics[width=0.8\linewidth]{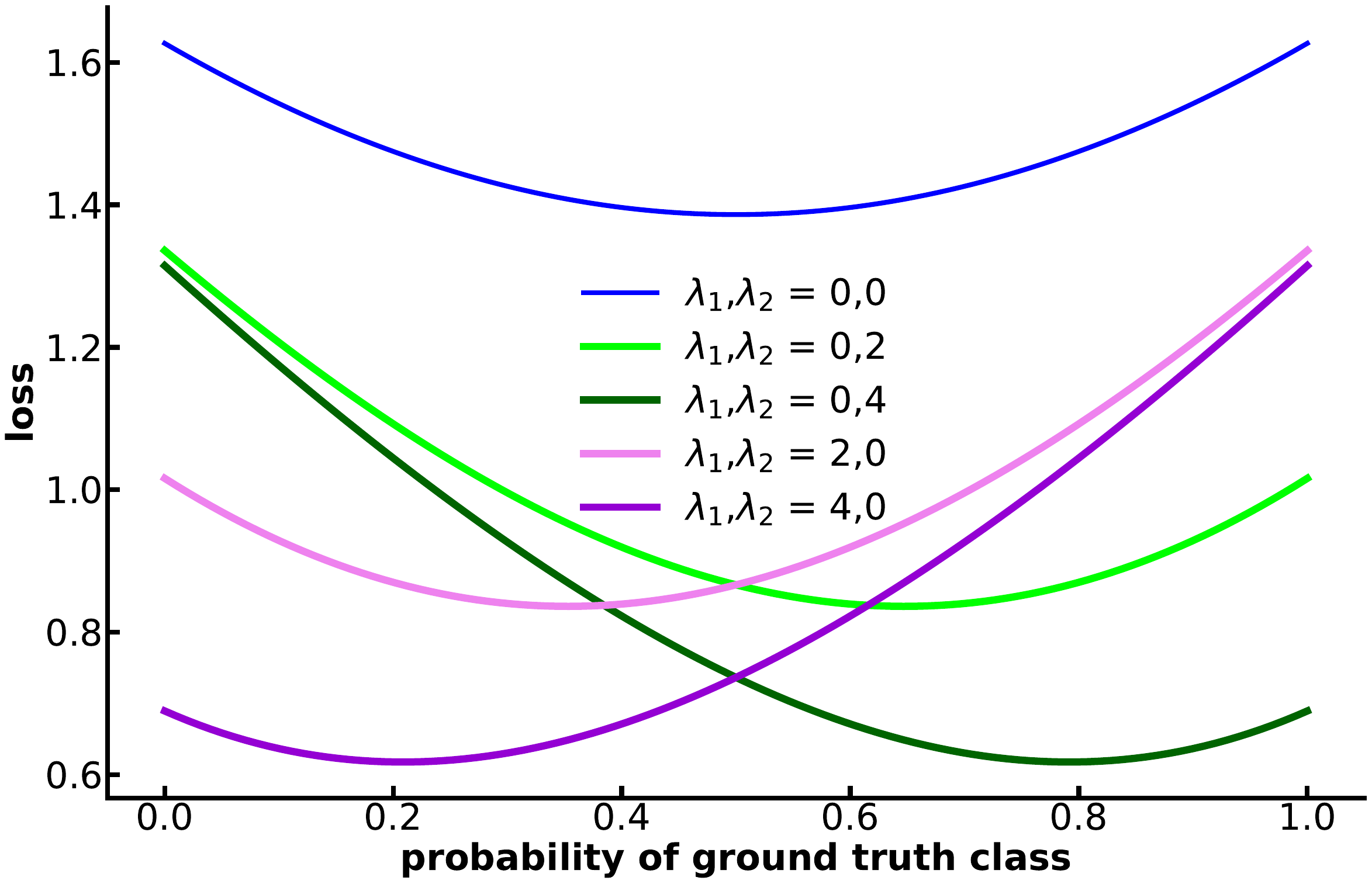}
    \caption{The comparison of the loss weighting factors for $L_{h}=-(1-p)^{\lambda _{1}}log(p) -(p)^{\lambda_{2}}log(1-p) $. }
    \label{fig:paremeter}
    \vspace{-6mm}
\end{figure}

}

Then the parameter $\varphi$ can be updated by minimizing the loss $L_{casl}$.

\section{Experiments}
The experiments contain the following four respects:
1) the analysis of two important parameters; 2) ablation studies in the
multi-view and re-weighting strategy;
3) comparison to the-state-of-the-art techniques; 
4) visualization of the sub-class performance by confusion matrix and the feature embedding by TSNE.
Our approach is mainly evaluated in the 3MDAD dataset~\cite{jegham2020novel}.
Before discussing the results, we provide the details of the experimental setup below.

\label{sec:typestyle}

\textbf{Dataset}
The 3MDAD dataset \cite{jegham2020novel} was captured by two Microsoft Kinect cameras installed on the car handle at the top of the passenger's window and the instrument panel in front of the driver respectively.

There are 50 drivers in the daytime where \{35, 5, 10\} drivers are randomly selected for \{train, val, test\}, respectively. 
All the frames were resized to 256 × 256.

\begin{figure}[t!]
    \centering
    \includegraphics[width=\linewidth]{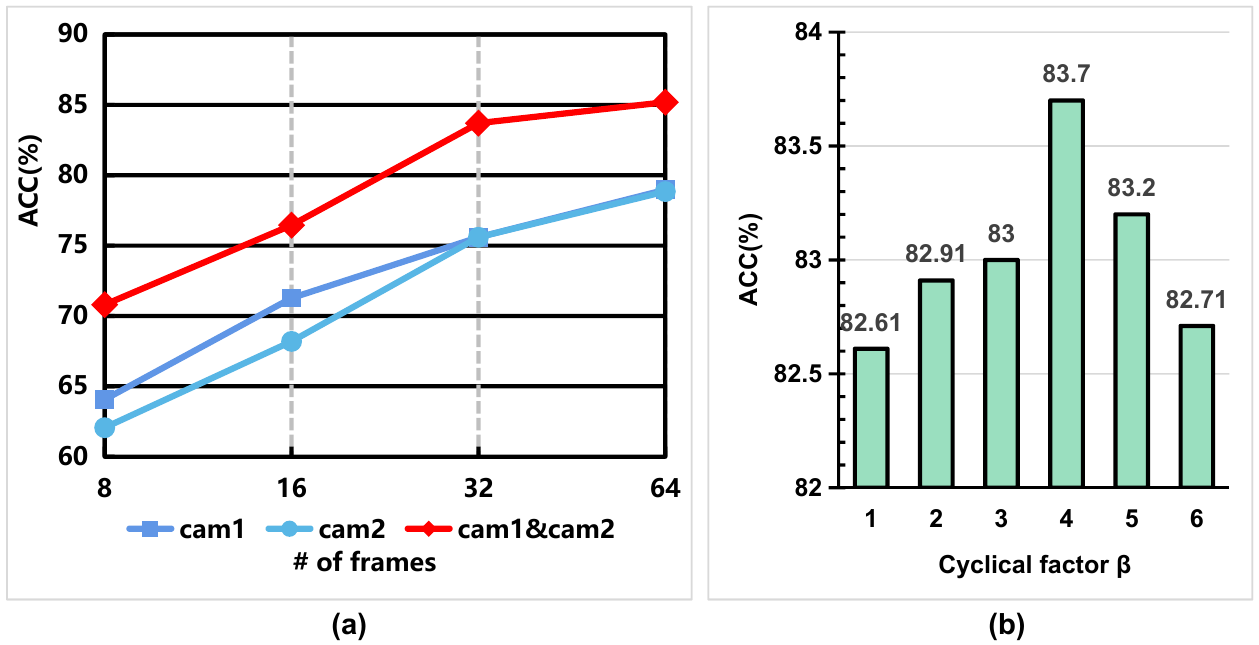}
    \caption{(a) The effect of the number of frames. (b) The effect of the cyclical factor $\beta$.}
    \label{fig:Nof-frame-cyclical-factor}
    \vspace{-6mm}
\end{figure}

\textbf{Experimental setting}
Our experiment is implemented in Python 3.6 development environment and based on the PyTorch 1.11 platform.
All models are trained on a server with Intel(R) Xeon(R) Gold 5218R CPU @ 2.10GHz, 188.00 GB memory, and 4 NVIDIA GeForce RTX 3090 GPUs.

For the single view backbone training, we take I3D as the backbone and use the pre-trained parameters on ImageNet as the initialization.
And we use Stochastic Gradient Descent (SGD) with an initial learning rate of 0.1 as the optimizer.
The models are trained for 1000 epochs where the learning rate is decayed at the [100th, 300th] epoch.
RandomRotation and RandomErasing \cite{zhong2020random} are applied for data augmentation.

For the multi-view fusion training, we take the pretrained single-view backbone as a feature extractor and freeze the weight of the backbones only updating the parameter from the fusion and re-weight part.
Similarly,  Stochastic Gradient Descent (SGD) with an initial learning rate of 0.1 is selected as the optimizer.
The model is trained for 100 epochs where the learning rate is decayed at the [30th, 50th] epoch.

\subsection{ Parameter Analysis} 
There are five parameters, number of frames,  $\beta$ (in Eq \ref{omega define}), $\gamma$ (in Eq \ref{Lp}), $\lambda _{1}$, $\lambda _{2}$  (in Eq \ref{Ln}) in our proposed MIFI model.

Following the setting in the literature \cite{smith2022cyclical}, we set the parameters $\gamma$=0, $\lambda _{1}$=0, $\lambda _{2}$=4.
Therefore, we mainly verify the influence of frame number and $\beta$ on the experimental results.
Additionally, we also discuss the influence of early fusion and later fusion.

\Lily{
\subsubsection{The effect of parameter $\lambda _{1}$ and $\lambda _{2}$}

It is important to note that choosing the right values for 
$\lambda _{1}$ and $\lambda _{2}$. If the $\lambda _{1}$ is too small, the model will focus too much on the hard samples and ignore easy samples, leading to poor performance on easy samples.
On the other hand, a too-small value of $\lambda _{2}$ 
will lead to the model achieving poor performance on hard samples.
Thus, we experiment with different values of $\lambda _{1}$ and $\lambda _{2}$ to find the optimal values for the proposed approach.
In this experiment, we use I3D as the backbone model.
We fix $\gamma = 0$, and we report the accuracy and F1-score under different parameter combinations on Table~\ref{Table:config}.
We can see that a too-small value of $\lambda _{2}$ leads to lower accuracy. And when we set a relatively small $\lambda _{1}$ and large $\lambda _{2}$, the accuracy and F1-score can be increased. This indicated that such a combination of $\lambda _{1}$ and $\lambda _{2}$ leads to 
the model can not only focus on hard samples but also not ignore the easy samples.
And it can be also observed that a too-large $\lambda _{2}$, such as  $\lambda _{2} = 6$, \textit{i,e,} almost neglecting learning features from easy samples,
the accuracy and F1-score declined.
This demonstrates the importance of the combination of the two terms in Eq.~\ref{Ln} in an appropriate way.
In the following experiments, we set $\lambda _{1}$ = 0 and large $\lambda _{2} = 4$.

\subsubsection{The influence of parameter $\gamma$}
 We conduct another experiment to study the impact of parameter $\gamma$. We also use I3D as the backbone model and fix $\lambda _{1}$ = 0 and large $\lambda _{2} = 4$.
Table~\ref{Table:gamma} shows the effect of parameter $\gamma$.
We can observe that $\gamma$  does affect the model performance as $\gamma$ is related to how much the easy sample will be weighted in the early training stage.
When $\gamma$ is set to a too-large value, like $6$, the accuracy is decreased.
And we can also see that when $\gamma$ is set in an appropriate range, the performance of the model is relatively stable.
In the following experiments, we set $\gamma=0$.

\begin{table*}[t!]
\large
    \caption{The effect of $\lambda _{1}$ and $\lambda _{2}$.}
    \centering
    \newcommand{\tabincell}[2]{\begin{tabular}{@{}#1@{}}#2\end{tabular}}
     \begin{tabular}{ c |c |c |c |c |c |c |c |c |c |c |c |c }
    \hline
    \hline
      $\lambda _{1}$  &0 &2 &4 &0 &2 &4 &0 &2 &4 &0 &2 &4 \\
    \hline
      $\lambda _{2}$  &0 &0 &0 &2 &2 &2 &4 &4 &4 &6 &6 &6 \\
    \hline
      ACC  &78.8 &76.4 &76.3 &76.8 &80.0 &77.4 &\textbf{83.7} &77.2 &77.8 &80.5 &76.9 &76.7\\
    \hline
    F1-score &78.6 &76.5 &76.6 &76.9 &78.7 &77.3 &\textbf{83.9} &77.3 &78.2 &81.2 &76.8 &76.6\\
    \hline
  \bottomrule
    \end{tabular}
    \label{Table:config}
\end{table*} 

\begin{table}[t!]
\large
    \caption{The effect of  $\gamma$.}
    \centering
    \newcommand{\tabincell}[2]{\begin{tabular}{@{}#1@{}}#2\end{tabular}}
     \begin{tabular}{ c |c |c |c |c }
    \hline
    \hline
      $\gamma$  &0 &2 &4 &6 \\
    \hline
      ACC  &\textbf{83.7} &79.6 &83.2 &79.64 \\
    \hline
    F1-score &\textbf{83.9} &79.4 &83.6 &79.8\\
    \hline
  \bottomrule
    \end{tabular}
    \label{Table:gamma}
    \vspace{-8mm}
\end{table} }

\subsubsection{The effect of the number of frames}

To verify the influence of the number of frames in each clip, we conduct experiments based on the single-view video, $i.e.,$ camera ID \#1 and camera ID \#2, and the proposed multi-view approach.
For all these three strategies, we take the I3D model as the backbone.
The influence of the number of frames is illustrated in Figure \ref{fig:Nof-frame-cyclical-factor} (a).
We can observe that: 
\Lily{
(1) The performance of the proposed method can be boosted with an increase in the number of frames, especially from 8 to 16. (2) Although the accuracy of 64 frames is increased in comparison to that of 32 frames, the increment from 32 frames to 64 frames becomes smaller compared to the increment from 8 frames to 16 frames.
The results suggest that increasing the number of frames can improve the performance of 3D models, but the improvement may diminish beyond a certain threshold. The choice of the number of frames should be based on the trade-off between performance and computational complexity. In this case, 32 frames are found to be sufficient for describing distraction behaviors while keeping the computational complexity manageable.
Thus, we choose to use 32 frames in each clip in the following experiments.
}

\subsubsection{The effect of cyclical factor $\beta$}
Cyclical factor $\beta$ determines the value of the periodic parameter $\alpha$  controlling the learning of the easy and hard samples.
Similarly, we take I3D as our backbone and observe the effect of the parameter $\beta$ on the proposed multi-view feature learning framework.
We vary the parameter $\beta$ from 1 to 6.
Technically, a small value of $\beta$ will encourage the model to focus more on the hard samples while a large value of  $\beta$ can enforce the model to pay more attention to the hard samples.
The experimental results are shown in Figure \ref{fig:Nof-frame-cyclical-factor} (b).
We can see that a too-small or too-large value of $\beta$ does influence the performance.
Fortunately, the accuracy is in a stable range [82.6, 83.7], this indicates the effectiveness of the proposed example re-weighting module.
In the following experiments, we set $\beta = 4$.

\begin{figure}[t!]
    \centering
    \includegraphics[width=\linewidth]{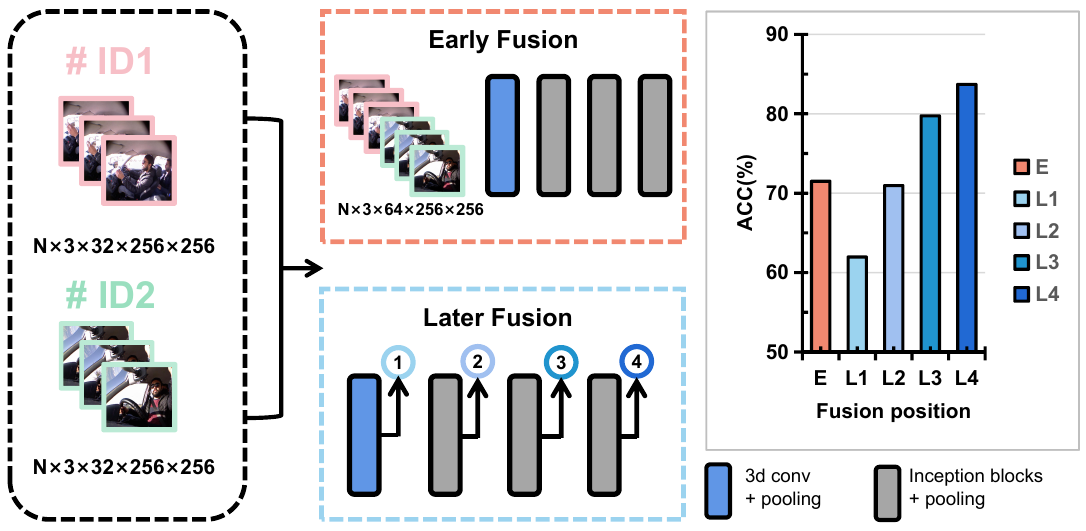}
    \caption{The effect of early fusion and later fusion. E represents early fusion, L represents the later fusion, and the number following the L represents the fusion position. }
    \label{fig:fusion position}
    \vspace{-6mm}
\end{figure}
\subsubsection{The effect of early fusion and later fusion}
To investigate the effect of early fusion and later fusion, take I3D as an example, we experiment with one early fusion method that directly fuses the inputs in the temporal dimension, and four types of later fusion approaches vary in the position that first extracts then fuses the extracted features, as shown in  Figure \ref{fig:fusion position}.
We can see that: 
(1) most later fusion methods can beat the early fusion model by a large margin. For example, L4 outperforms E by more than 10\%.
This demonstrates that later fusion models are more effective than early fusion models.
(2) the results from L1 to L4 indicate that the later position, the better the fusion performance.

\Lily{
(3) Interestingly, it can be noted that the performance of $L1$ that fuses the shallow features is the lowest even compared with the early fusion (E) that directly uses the image-level feature.
We hypothesize that when multiple inputs capture the same driver against a similar background, the resulting shallow features from multiple views may be too similar and unable to complement each other at the feature level. This is due to the fact that the features generated from the shallow layer primarily represent color, edges, and texture information.
On the other hand, we can also see that at the feature level, the later the fusion, the better the performance.

}

\subsection{Ablation Studies}
There are mainly two modules in the proposed MIFI model, \textit{i.e.,} multi-view feature fusion and example re-weighting.
Therefore, in this section, we verify their necessity by an ablation study.
We also take I3D as our feature extractor and 3MDAD as our experimental dataset.

\begin{figure*}[t!]
\centering
	\subfloat[Camera\#1 ($L_{CE}$)]{\includegraphics[height=0.23\linewidth]{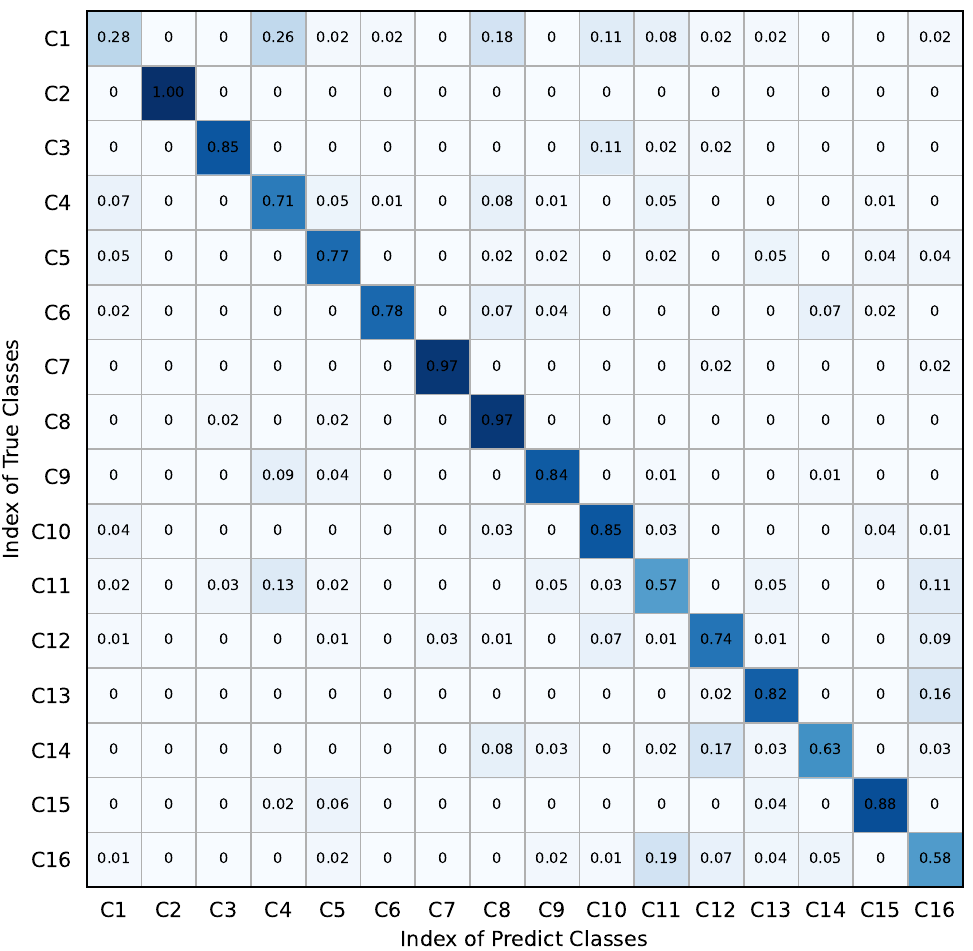}}
	\hfill
	\subfloat[Camera\#2 ($L_{CE}$)]{\includegraphics[height=0.23\linewidth]{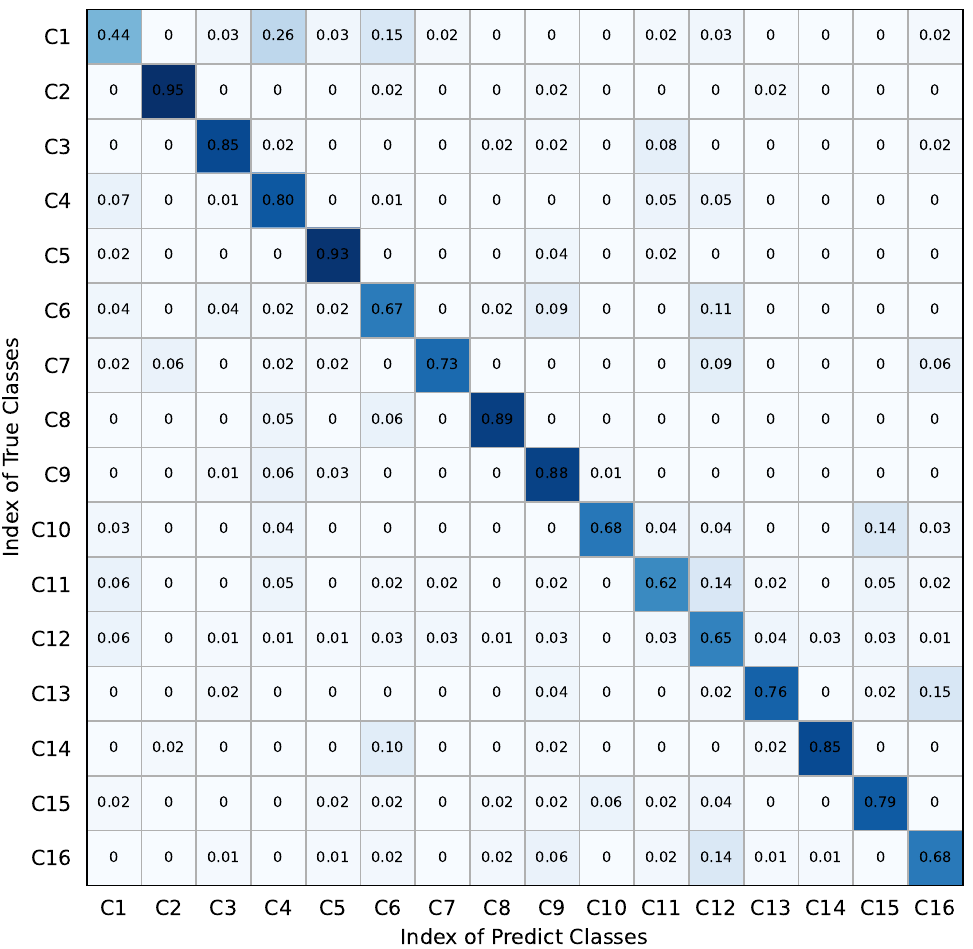}}
	\hfill
	\subfloat[Two-view ($L_{CE}$)]{\includegraphics[height=0.23\linewidth]{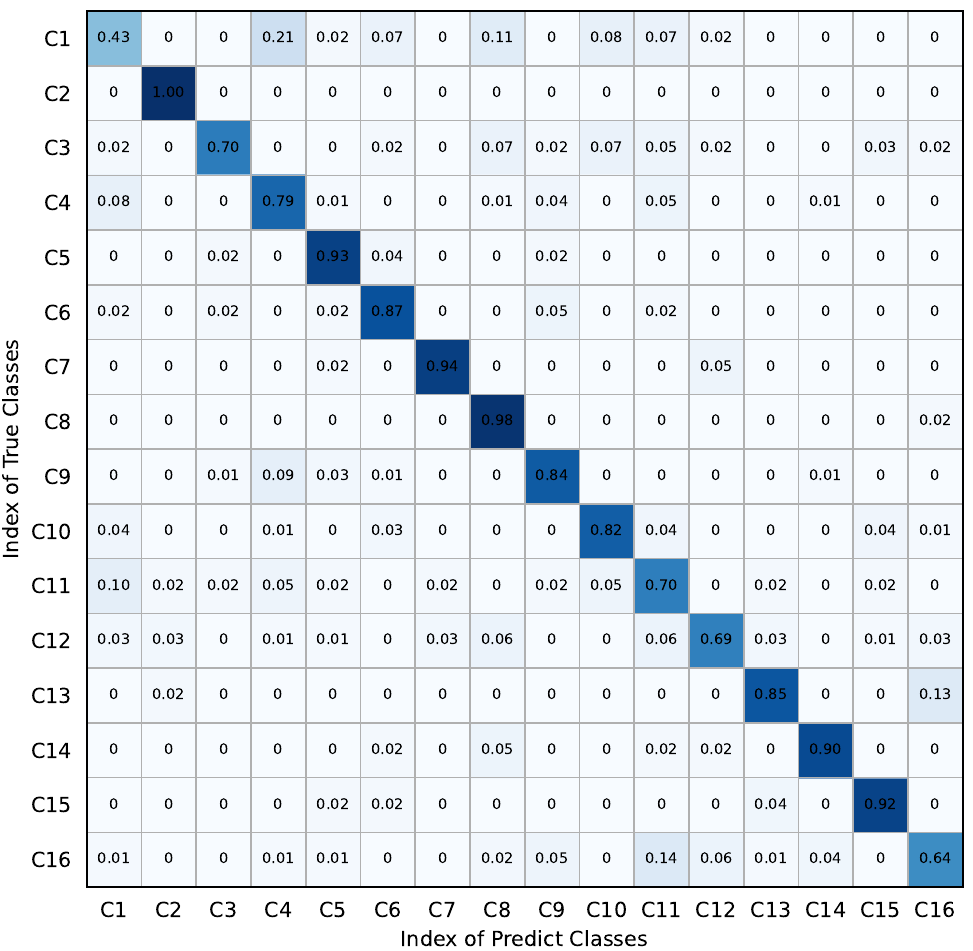}} 
	\hfill
	\subfloat[Two-view ($L_{CASL}$)]{\includegraphics[height=0.23\linewidth]{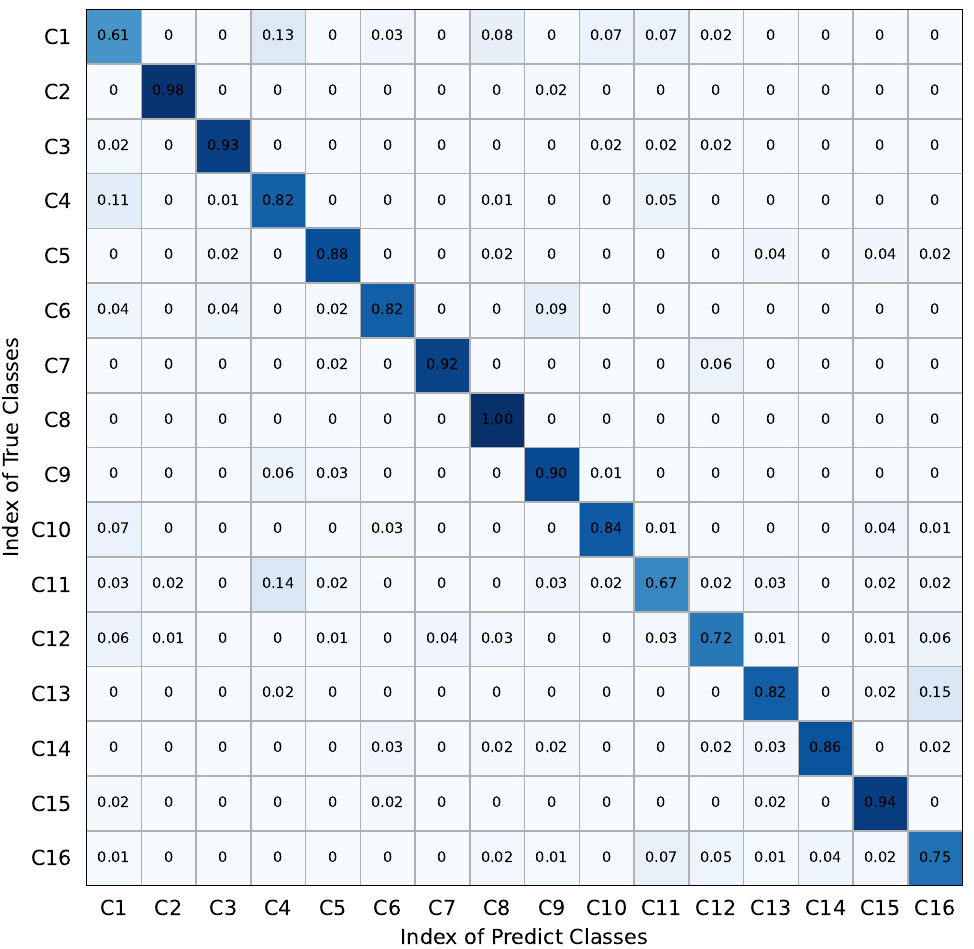}} 
	\hfill
	\subfloat{\includegraphics[height=0.23\linewidth]{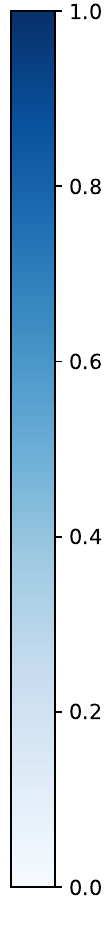}} 
\caption{Confusion matrix of the single-view and multi-view models. The best view is in zoom.}
\label{fig:matrix}
\vspace{-4mm}
\end{figure*}


\begin{figure*}[t!]
\centering
	\subfloat[Camera\#1 ($L_{CE}$)]{\includegraphics[width=0.24\linewidth]{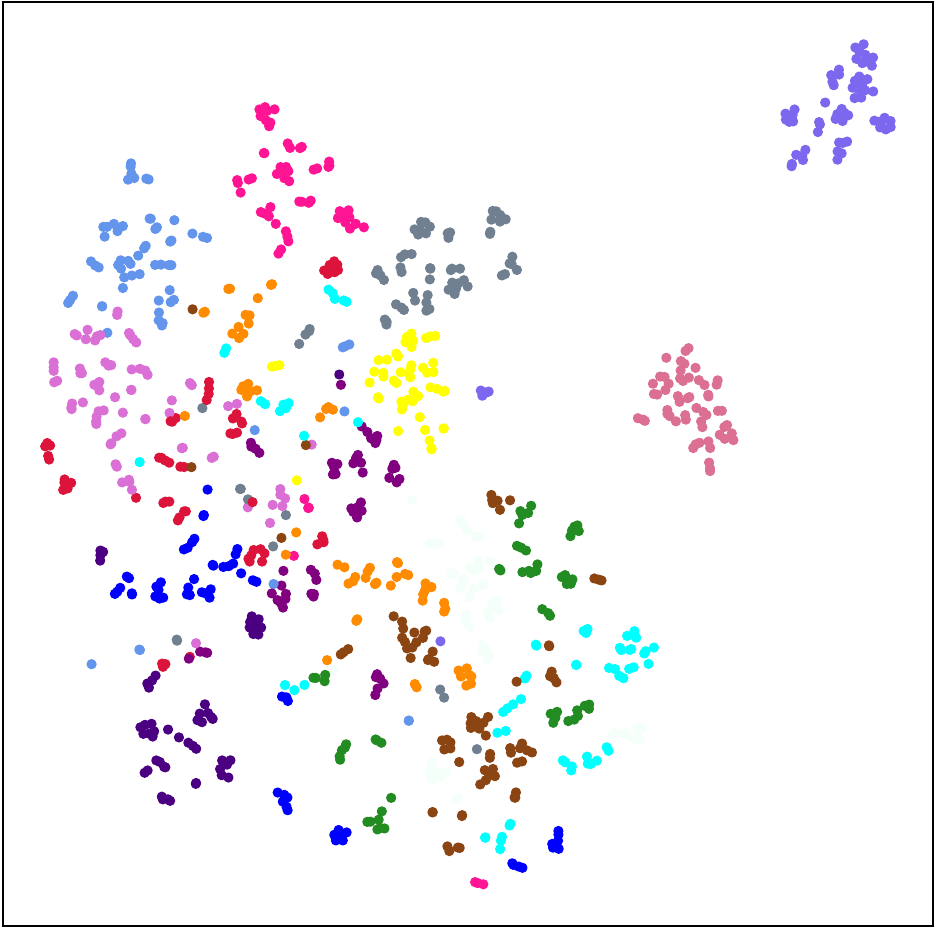}}
	\hfill
	\subfloat[Camera\#2 ($L_{CE}$)]{\includegraphics[width=0.24\linewidth]{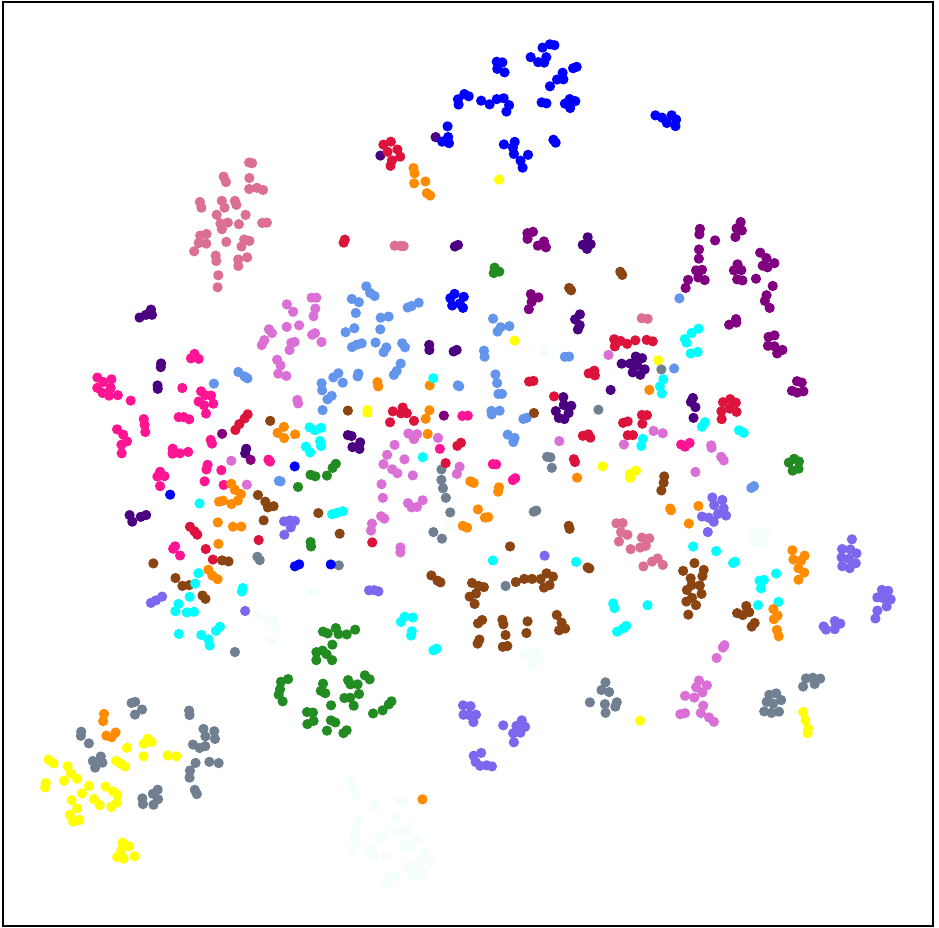}}
	\hfill
	\subfloat[Two-view ($L_{CE}$)]{\includegraphics[width=0.24\linewidth]{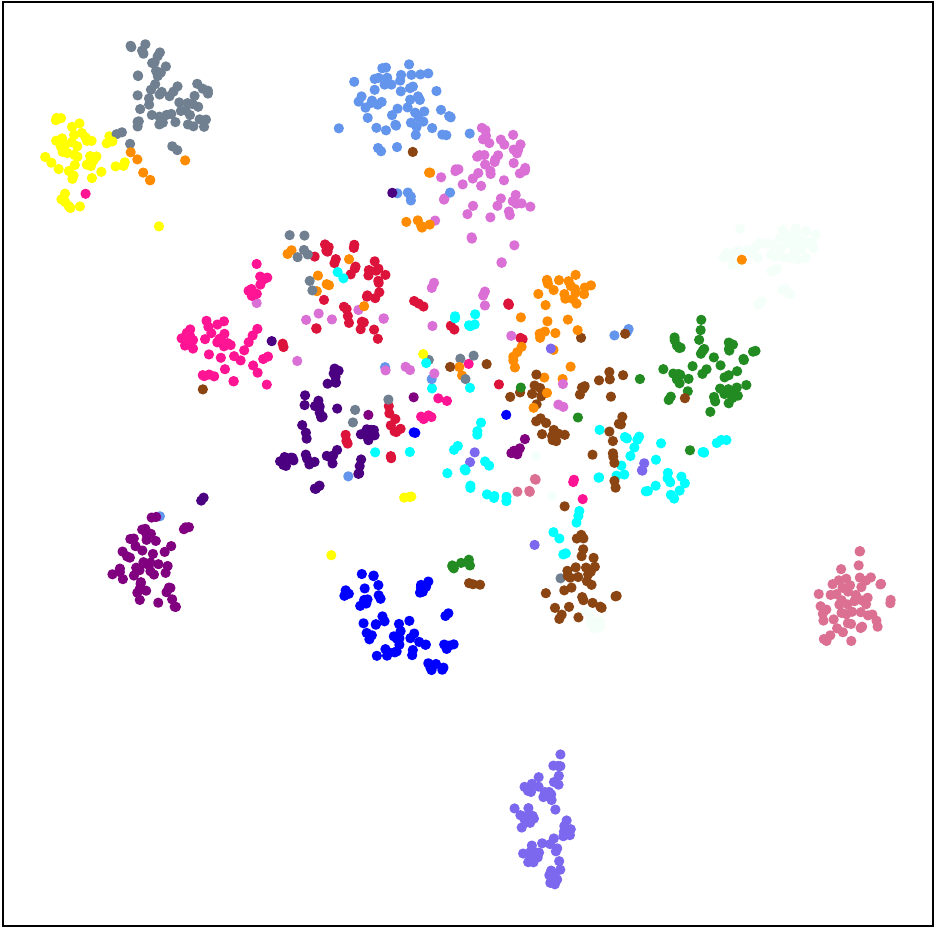}} 
	\hfill
	\subfloat[Two-view ($L_{CASL}$)]{\includegraphics[width=0.24\linewidth]{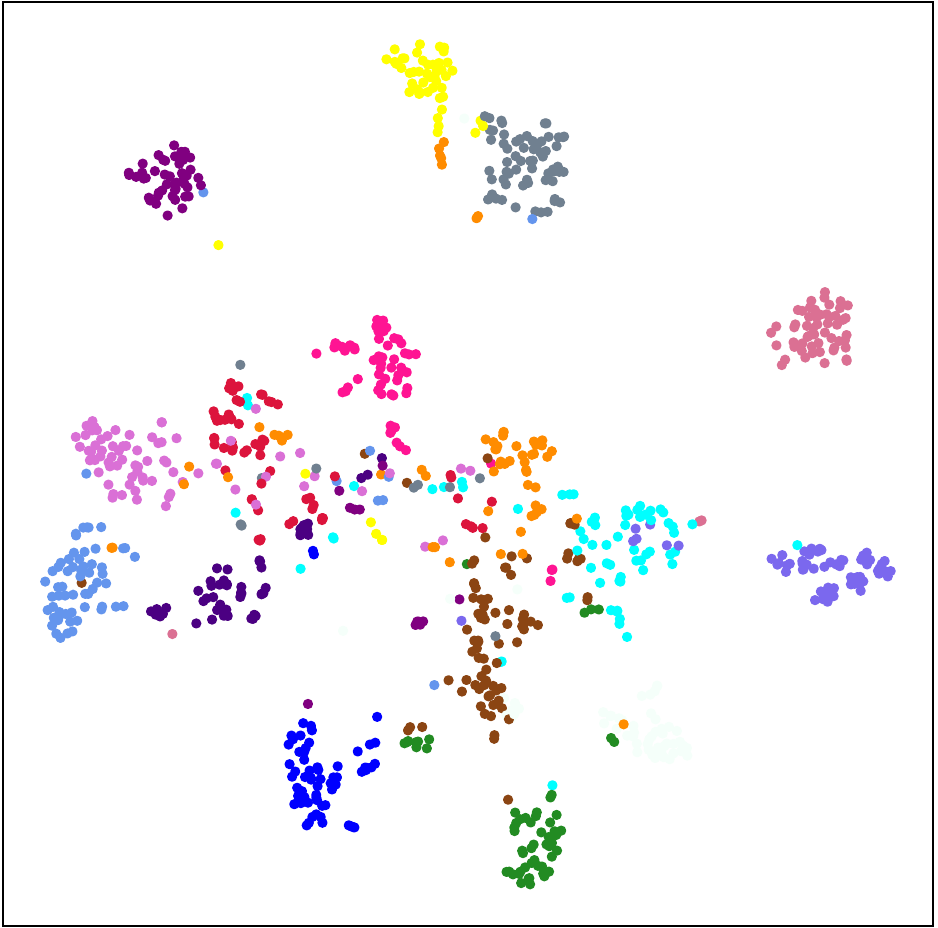}} 
	\hfill
	\subfloat{\includegraphics[width=\linewidth]{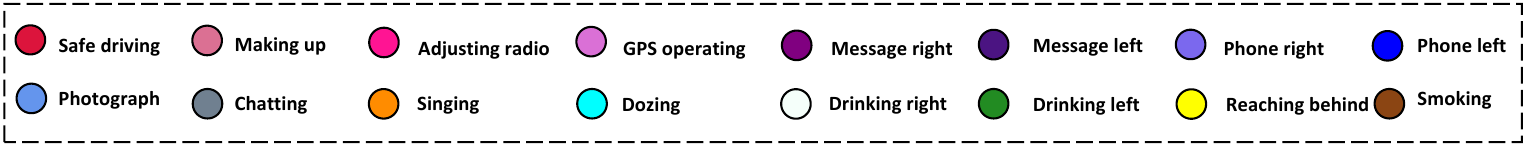}} 
\caption{TSNE visualization of the single-view and multi-view models. The best view is in zoom.}
\label{fig:tsne}
\end{figure*}

\textbf{Multi-view Fusion}
In order to evaluate the effectiveness of the multi-view feature learning module,
we perform a single-view version of the backbone and a MIFI-based two-view fusion method.
To further verify the robustness of the proposed MIFI, we also vary the number of frames in each clip.
For all the models, we take the cyclical focal loss to update the parameters.
The results are listed in Table \ref{Table:fusion}.
We can observe the clear improvements of the proposed MIFI no matter how many frames.
For instance, when the number of frames is 32, the two-view fusion model gains 5.04\% in accuracy.
This indicates that the proposed multi-view feature integration is more powerful for learning feature representation.

\begin{table}[t!]
\normalsize
    \caption{The accuracy of single-view and multi-view I3D under different frame numbers. $\textcolor{green}{\uparrow}$ represents the increased accuracy compared to the best results of the single-view model. Note that all the models are based on $L_{CASL}$ loss.}
    \centering
    \newcommand{\tabincell}[2]{\begin{tabular}{@{}#1@{}}#2\end{tabular}}
    \begin{tabular}{c|c|c|c}
    \toprule
   \# Frames & Camera \#1 & Camera \#2 & Two-view   \\
  \hline
  \hline
      8  &64.1 &62.1 &68.6 ($\textcolor{green}{\uparrow}$ \textbf{4.5})\\
      16 &71.3 &68.2 &74.8 ($\textcolor{green}{\uparrow}$ \textbf{3.5})\\
      32 &75.6 &75.6 &80.6 ($\textcolor{green}{\uparrow}$ \textbf{5.0})\\
      64 &79.0 &78.8 &82.7 ($\textcolor{green}{\uparrow}$ \textbf{3.7})\\
  \bottomrule
    \end{tabular}
    \label{Table:fusion}
    \vspace{-6mm}
\end{table}


\begin{table*}[h]

\normalsize
    \caption{The experimental results on the 3MDAD dataset. TSM-Res: TSM-Resnet101, TSM-Mo: TSM-Mobilenetv2, TSM-BN: TSM-BNInception. Concat(C) and Concat(T) represent the concatenation in channel and temporal dimensions, respectively.
    Para: Parameter size, F1: F1-score, F1-1: F1-score of Camera \#1, F1-2: F1-score of Camera \#2.}
    \centering
     \setlength{\tabcolsep}{1.0pt}

    \begin{tabular}{c|ccccc|cccc|cccc|cccc|cccc}
    \toprule
   \multirow{2}{*}{\textbf{Model}} & \multicolumn{5}{c|}{Single-view} & \multicolumn{4}{c|}{Voting} & \multicolumn{4}{c|}{MIFI (Sum)}& \multicolumn{4}{c|}{MIFI (Concat-C)} & \multicolumn{4}{c}{MIFI (Concat-T)} \\
   
    & Para & FLOPs & FPS & F1-1 & F1-2  
    & Para & FLOPs & FPS & F1  
    & Para & FLOPs & FPS & F1  
    & Para & FLOPs & FPS & F1  
    & Para & FLOPs & FPS & F1  \\ 

    \hline
 \textbf{ResNet50} & 23.5 & 8.3 & 66.9 & 51.9 & 49.7  & 47.0 & 16.6 & 34.0 & 57.9 & - & - & - & - & - & - & - & - & - & - & - & - \\
    \hline
\textbf{R(2+1)d}   &27.3  & 233.3 & 28.1 & 70.6 & 75.0  & 54.6 &466.6 & 15.8 & 78.6 & 54.6 & 466.6  & 15.8 & 78.8 & 54.6 & 466.6 & 16.5 & 79.8   & 54.6 & 466.6 &16.3 & 80.5 \\
\textbf{TSM-Res}   & 42.5 & 503.3 & 21.8 & 63.8 & 66.9 &85.1 &1006.7 &11.4 &70.9
& 85.0 &1006.7 & 11.1 & 68.8  & 85.1 &1006.7 & 11.2 & 68.8 & 85.1 & 1006.7 & 11.0 & 70.8 \\
\textbf{TSM-Mo }  & 2.2 & 20.9 &38.3 & 55.9 &52.8 &4.5 &41.8 &22.3 &61.1 &4.5 &41.8 & 22.8 &65.8  & 4.5 & 41.8 & 22.6 & 64.7  &4.5 &41.8 &22.8 & 63.0 \\
\textbf{TSM-Bn}  & 10.3	& 131.9 & 27.3 & 62.2 & 61.0 &20.6 & 262.1 &13.5 & 66.5 &20.6 & 262.1 & 13.6 & 65.7  &20.6 & 262.1 & 13.4 & 69.4 &20.6 & 262.1 & 14.2 & 66.6 \\
\textbf{X3D}   & 3.0 & 20.3 & 22.5 & 61.5 & 65.0 &6.0 &40.5 &11.9 &76 & 4.0 & 39.9 & 12.4 & 72.4  & 4.0 & 39.9 &13.0 & 76.8  & 4.0 &39.9 &12.9 & 73.8 \\
\textbf{I3D}   & 12.3 & 111.5 & 25.8 & 75.8 & 75.6  & 24.6 & 223.0 & 14.1 &81.3 & 24.6 & 223.0 & 14.7 & 82.5 & 24.6 & 223.0 & 14.7 & 82.0 & 24.6 &223.0 &14.9 & 83.9 \\
   \bottomrule
    \end{tabular}
    \label{table:para-macs-acc-R1}
    \vspace{-4mm}
\end{table*}

\textbf{Example Re-weighting}
To verify the effectiveness of the proposed example re-weighting module, we conduct experiments based on I3D with different example re-weighting loss, including Focal Loss ($L_{FL}$), Asymmetric Loss ($L_{ASL}$), and the utilized  Cyclical Focal Loss($L_{CASL}$).
And we also conduct Cross-entropy Loss ($L_{CE}$) as the baseline.
Also, single-view and multi-view frameworks are used.
The results are shown in Table \ref{Table:reweight}.
It can be seen that: (1) By comparing the
results of $L_{CE}$ and $L_{CASL}$, find that the proposed example re-weighting module can consistently contribute to the accuracy on both single-view and two-view model.
This demonstrates the effectiveness of the proposed example re-weighting module.
(2) By comparing the
results of $L_{CE}$ and $L_{FL}$, $L_{ASL}$, we can observe that the latter two losses are not able to improve the performance in most cases.
This is because the data distribution in 3MDAD is relatively balanced whilst such losses are for imbalanced data, which further indicates that inappropriate example re-weighting models may hurt the performance.
(3) Surprisingly, the $L_{FL}$ based on the two-view framework gains 0.5\% improvement, this also verifies the benefit of the multi-view framework.

To show the performance in the sub-classes, we 
draw the confusion matrix of the two $L_{CE}$ based single-view models, $L_{CE}$ and $L_{CASL}$ based multi-view methods on Figure \ref{fig:matrix}.
First, it can be observed that the difficulty inconsistent phenomenon exists, for example, the class C1 (\textit{Safe driving}), C11(\textit{Singing}), C12 (\textit{Dozing}), and C16 (\textit{Smoking}) are much harder than the other classes, in all these four frameworks, which indicates that the example re-weighting is pretty necessary.
Second, compared the matrix of Camera \#1 ($L_{CE}$), Camera \#2 ($L_{CE}$), Two-view ($L_{CE}$), the color of the blocks on the diagonal in the Two-view ($L_{CASL}$) is relatively close, $\textit{i.e.,}$ the accuracy of the subclasses is relatively close.
This verifies that the example re-weighting combined with the two-view framework can help the model to learn more appropriate parameters that are beneficial for the classification.

To further investigate the difference between the learned features in a simple and clear way, we project the learned features into 2D space by TSNE\cite{JMLR:v9:vandermaaten08a}.
From Figure \ref{fig:tsne}, we can see that:
(1) Compared to the results of Two-view ($L_{CE}$), the learned features from the  Two-view ($L_{CASL}$), are more clustered, \textit{i.e.,} the intra-class variation is small, and the inter-class variation is large.
This verifies the effectiveness of the proposed example re-weighting module that is able to encourage the model to learn more efficient feature representation.
(2) Compared to the results of Camera\#1 ($L_{CE}$), Camera\#2 ($L_{CE}$), the learned features from the  Two-view ($L_{CE}$) seems more compact, which shows the effect of the proposed multi-view feature integration framework.


\begin{table}[t!]
\normalsize
    \caption{The recognition accuracy of different re-weighting models.\{$L_{CE}$, $L_{FL}$, $L_{ASL}$, $L_{CASL}$ \} are the \{Cross-entropy, Focal,  Asymmetric,  Cyclical Focal\} Loss, respectively. $\textcolor{green}{\uparrow}$ and $\textcolor{red}{\downarrow}$ represents the increased and decreased accuracy compared to the $L_{CE}$ loss.}
    \centering
    \newcommand{\tabincell}[2]{\begin{tabular}{@{}#1@{}}#2\end{tabular}}
     \begin{tabular}{ c| c |c |c}
      \hline
       Loss & Camera \#1 & Camera \#2 & Two-view \\
      \hline
      \hline
     $L_{CE}$  &75.6 &75.6 &80.6\\
      $L_{FL}$ & 68.3 ($\textcolor{red}{\downarrow}$7.3) &73.4 ($\textcolor{red}{\downarrow}$2.2) &81.1 ($\textcolor{green}{\uparrow}$0.5)\\
     $L_{ASL}$  &68.8 ($\textcolor{red}{\downarrow}$ 6.8) &74.4($\textcolor{red}{\downarrow}$ 1.2) &80.1 ($\textcolor{red}{\downarrow}$ 0.5)\\
     $L_{CASL}$ &78.2 ($\textcolor{green}{\uparrow}$ \textbf{2.6}) &78.8 ($\textcolor{green}{\uparrow}$ \textbf{3.2}) & 83.7 ($\textcolor{green}{\uparrow}$ \textbf{3.1}) \\
  \bottomrule
    \end{tabular}
     \label{Table:reweight}
     \vspace{-8mm}
\end{table}

\subsection{Compare with the state-of-the-art}
In this section, we compare our method with state-of-the-art 3D action recognition methods.
For a fair comparison, we use the same configuration for all models.
To verify the effectiveness of the proposed MIFI, we also implement the multi-view version of all the methods in which we keep the same feature extractor layer and add the multi-camera feature fusion and example re-weighting modules in such methods.

\subsubsection{Backbones}

a) R(2+1)d \cite{tran2018closer}: demonstrate the accuracy advantages of 3D CNNs over 2D CNNs within the framework of residual learning and show that factorizing the 3D convolutional filters into separate spatial and temporal components yields significant gains in accuracy.

b) I3D \cite{carreira2017quo}: learn seamless spatio-temporal feature extractors from video by expanding filters and pooling kernels of very deep image classification ConvNets into 3D while leveraging successful ImageNet architecture designs and even their parameters.

c) TSM \cite{lin2019tsm}: propose a generic
and effective Temporal Shift Module that enjoys both
high efficiency and high performance. TSM facilitates information exchanged among
neighboring frames in the way of shifting part of the channels along the temporal dimension.

d) X3D \cite{feichtenhofer2020x3d}: a family of efficient video networks that progressively expand a tiny 2D image classification architecture along multiple network axes, in space, time, width, and depth.

\Lily{
e) Voting:  a simple voting multi-camera method that directly selects the class with the highest probability of the two views as the prediction.

f) ResNet50: we also compare image-based DDC based on ResNet50.
Note that we use the same training data set for the image-based DDC and video-based DDC.
And we follow the same training setting with the work~\cite{li2022learning}.

}

\subsubsection{Analysis}
From the table~\ref{table:para-macs-acc-R1} we can obtain the following observations.

\Lily{
$(\romannumeral1)$ The proposed MIFI can consistently improve performance under various backbone models.
For example, for I3D, MIFI (Sum), MIFI (Concat-C), and MIFI (Concat-T) beat the best single-view model by +6.7\%, +6.2\%, +8.1\%, respectively.
This indicates the superiority of the proposed MIFI framework.

$(\romannumeral2)$ MIFI (Concat-T) receives the best accuracy gaining +8.1\% improvement.
And we can also find that the MIFI (Concat-T) outperforms the other two methods, especially MIFI (ConcatC),
which suggests that temporal information is crucial for recognizing distracted driving behaviors and that concatenating features over time can lead to better performance than concatenating features across channels.
Furthermore, it can be seen that the MIFI (Sum) method achieves the lowest accuracy compared to MIFI (Concat-T) and MIFI (ConcatC), which also imply that directly summarizing the features from different views may not be an effective way to capture the discriminative information in the data.

$(\romannumeral3)$ The fusion performance relies on the backbone feature extractor. For example,
in comparison to the other backbones, 
I3D shows the superiority in distracted driver recognition tasks which can achieve the best accuracy.
This tells us that the following research
can directly take the I3D as the backbone to extract the features in the DDC task.

$(\romannumeral4)$ Although the image-based DDC, \textit{i,e,} ResNet50 requires fewer computational resources (see FLOPs), the inference accuracy is far from satisfactory.
For example, in the single-view scenario, the I3D can outperform ResNet50 by almost 25\% under all the camera views.
This indicates that the video-based DDC can be a better choice for the accuracy-priority DDC task.

$(\romannumeral5)$ The parameter size and computational complexity of multi-view models will be increased linearly, which is an accuracy-speed trade-off problem.
For this problem, there are two directions.
On one hand, we can design more efficient backbones that are with less parameter size and computational load.
On the other hand, we can consider implementing parallel processing on hardware\cite{zhou2022real} that can process the input from different cameras simultaneously.

$(\romannumeral6)$ We can see that simply voting on the two predictions can improve the performance compared to the single-view method.
But compared with the proposed MIFI, especially, MIFI (Contact-T), the improvement is limited under the similar computational cost.
This indicates that the proposed MIFI can integrate features from multiple views in a more efficient way.
}

\subsection{Further Analysis}
\Lily{
One weakness of 3D CNNs is that they require relatively high
computational costs and more training samples.
To reduce the computational cost, Key frame extraction that 
selects fewer keyframes to represent the input clip may be able to reduce the model complexity.
Thus, here, we study the effect of the key frame extraction approach and the amount of training data.

\subsubsection{Key Frame Extraction}
We incorporate the key frame extraction method (KFE) \cite{cheng2014motion} into the proposed 
approach. 
The KFE module first computes the pairwise differences between the frames and then selects the top $n$ frames that have the largest difference as the keyframes.
We also compared the results of single-view and multi-view implementation. 
We apply the I3D as the backbone model.
We take accuracy and clips per second (CPS) as our evaluation metric and the results are listed in Table~\ref{table:key-frame}.
First, by comparing the accuracy of single-view and two-view methods, the benefit of multi-view fusion is significant no matter whether the input frame size is 8 or 32.
Second, it can be seen that although the complexity of the model with the KFE module is reduced, the accuracy and inference speed of the model with KFE (32$\to$8) is dramatically decreased in comparison to the model directly using 32 frames as input.
This is because the introduction of key frame extraction brings additional computational overhead. 
Thus, it is critical to design a KFE method that can not only effectively extract the keyframes of the input but also does not introduce extra overhead.

\begin{table}[!t]
\normalsize
    \caption{The comparison of the models with and without key frame extraction approach. KFE: key frame extraction. CPS: clips per second.}
    \centering
     \setlength{\tabcolsep}{1.0pt}

    \begin{tabular}{c|cccc|ccc}
    \toprule
   \multirow{2}{*}{\# Frames} & \multicolumn{4}{c|}{Single-view} & \multicolumn{3}{c}{Two-view}  \\
   
    & CAM1 & CAM2  &CPS &FLOPs
    & ACC  &CPS  &FLOPs\\ 

    \hline
    \hline
 \textbf{8} & 64.1 & 62.1 & 27.9 &27.9 & 68.6 & 15.3 &55.8\\
\textbf{32}   &75.6  & 75.6 & 25.8 &111.5 & 80.6 & 14.9 &222.5\\
\textbf{KFE (32$\rightarrow$8)}   & 67.9 & 66.5 & 5.5  &27.9 & 71.6 & 2.1 &55.8 \\
   \bottomrule
    \end{tabular}
    \label{table:key-frame}
\end{table}

\subsubsection{The influence of the volume of the training data}

We gradually decreased the number of drivers used for training, and the results are listed in Table~\ref{Table:peo}. It can be seen that the performance of all models decreased with the decrease in the training data volume. This is due to the fact that 3D-based models require adequate data for training, which is a general problem in deep learning models. Fortunately, the proposed MIFI model still retains its advantages. Regardless of the amount of training data, the MIFI can improve performance. This further suggests the superiority of the proposed multi-camera feature integration.

    \begin{table}[t!]
\normalsize
    \caption{The experimental results on the 3MDAD dataset. Concat(T) represents the concatenation in the temporal dimension, respectively. \# Drivers: \# Drivers for Training. }
    \centering
    \newcommand{\tabincell}[2]{\begin{tabular}{@{}#1@{}}#2\end{tabular}}
     \begin{tabular}{ c |c |c |c }
      \hline
       \# Drivers & Camera \#1 & Camera \#2 & MIFI (Concat-T)  \\
      \hline
      \hline
      15  &56.4 &58.2 &59.4 \\
      25  &64 &66.3& 72.4\\
      35  &75.8 &75.6 &83.9 \\
  \bottomrule
    \end{tabular}
    \label{Table:peo}
    \vspace{-6mm}
\end{table} 
}

\section{Conclusion}
In this paper, we propose a new multi-camera feature integration, called MIFI, for robust 3D distracted driver activity recognition.
To effectively fuse the feature from multiple videos, three types of feature fusion approaches are presented.
To address the difficulty inconsistent problem in DDC, a simple yet effective example model is proposed.
By combining the two modules, the proposed MIFI can consistently boost the performance of current 3D action recognition models.
Additionally, our MIFI framework also can be utilized for other same-modality multi-view tasks, such as human action recognition, person re-identification, and \textit{etc.}

However, there remain some issues.
First, our goal in this work is to improve the accuracy of the model. The parameter size and computation complexity of the model are relatively less considered.
Second, the proposed method improves the effect of driving behavior recognition on the basis of existing models. 
Second, the proposed MIFI model is designed for recognizing driver distractions in the daytime, while the generalization ability to nighttime is not studied.
In future work, we will focus on the next directions. 
(1) We will explore a light-weighting version of 3D models for the DDC task to improve the 
meet the requirement of real-world scenarios.
(2) We will further investigate the model in a more practical environment such as nighttime, thus improving the model generalization ability.


\bibliographystyle{IEEEtran}
\bibliography{paper}

\end{document}